\documentclass[10pt,twocolumn,letterpaper]{article}

\usepackage{cvpr}              

\usepackage[dvipsnames]{xcolor}

\definecolor{cvprblue}{rgb}{0.21,0.49,0.74}
\usepackage[pagebackref,breaklinks,colorlinks,citecolor=cvprblue]{hyperref}

\def\paperID{*****} 
\def\confName{CVPR}
\def\confYear{2024}

\title{CSDN: A Context-Gated Self-Adaptive Detection Network for Real-Time Object Detection}

\author{Wei Haolin
}

\begin{document}
\maketitle
\begin{abstract}
Convolutional neural networks (CNNs) have long been the cornerstone of target detection, but they are often limited by limited receptive fields, which hinders their ability to capture global contextual information. We re-examined the DETR-inspired detection head and found substantial redundancy in its self-attention module. To solve these problems, we introduced the Context-Gated Scale-Adaptive Detection Network (CSDN), a Transformer-based detection header inspired by human visual perception: when observing an object, we always concentrate on one site, perceive the surrounding environment, and glance around the object. This mechanism enables each region of interest (ROI) to adaptively select and combine feature dimensions and scale information from different patterns. CSDN provides more powerful global context modeling capabilities and can better adapt to objects of different sizes and structures. Our proposed detection head can directly replace the native heads of various CNN-based detectors, and only a few rounds of fine-tuning on the pre-trained weights can significantly improve the detection accuracy.
\end{abstract}    
\section{Introduction}
\label{sec:introduction}

Object detection remains a pivotal task in computer vision, underpinning diverse advanced applications from autonomous driving and medical imaging to security surveillance and robotics. For years, convolutional neural networks (CNNs), particularly architectures like the YOLO series \cite{Redmon2016CVPR, Bochkovskiy2020arXiv}, have been the cornerstone of object detection due to their inherent efficiency and speed, making them ideal for real-time scenarios. However, CNNs inherently suffer from limited receptive fields, struggling to capture sufficient global contextual information, especially with varying object scales or occlusions. While architectural improvements like Feature Pyramid Networks (FPNs) \cite{Lin2017FPN} mitigate some scale issues, their capacity for robust long-range dependency modeling remains constrained.


The emergence of Transformer architectures \cite{Vaswani2017NeurIPS}, initially successful in Natural Language Processing, promised a solution to this limitation by enabling global information aggregation through attention mechanisms. This led to the development of Transformer-based detectors like DETR \cite{Carion2020ECCV}, which can model long-range dependencies more effectively, often yielding higher accuracy. However, these early models were typically computationally intensive and suffered from slow inference speeds, prompting a recent trend of hybrid approaches that integrate Transformer components, often into the detection head, to balance accuracy and efficiency.

\begin{figure}[h!]
\centering
\includegraphics[width=0.25\textwidth]{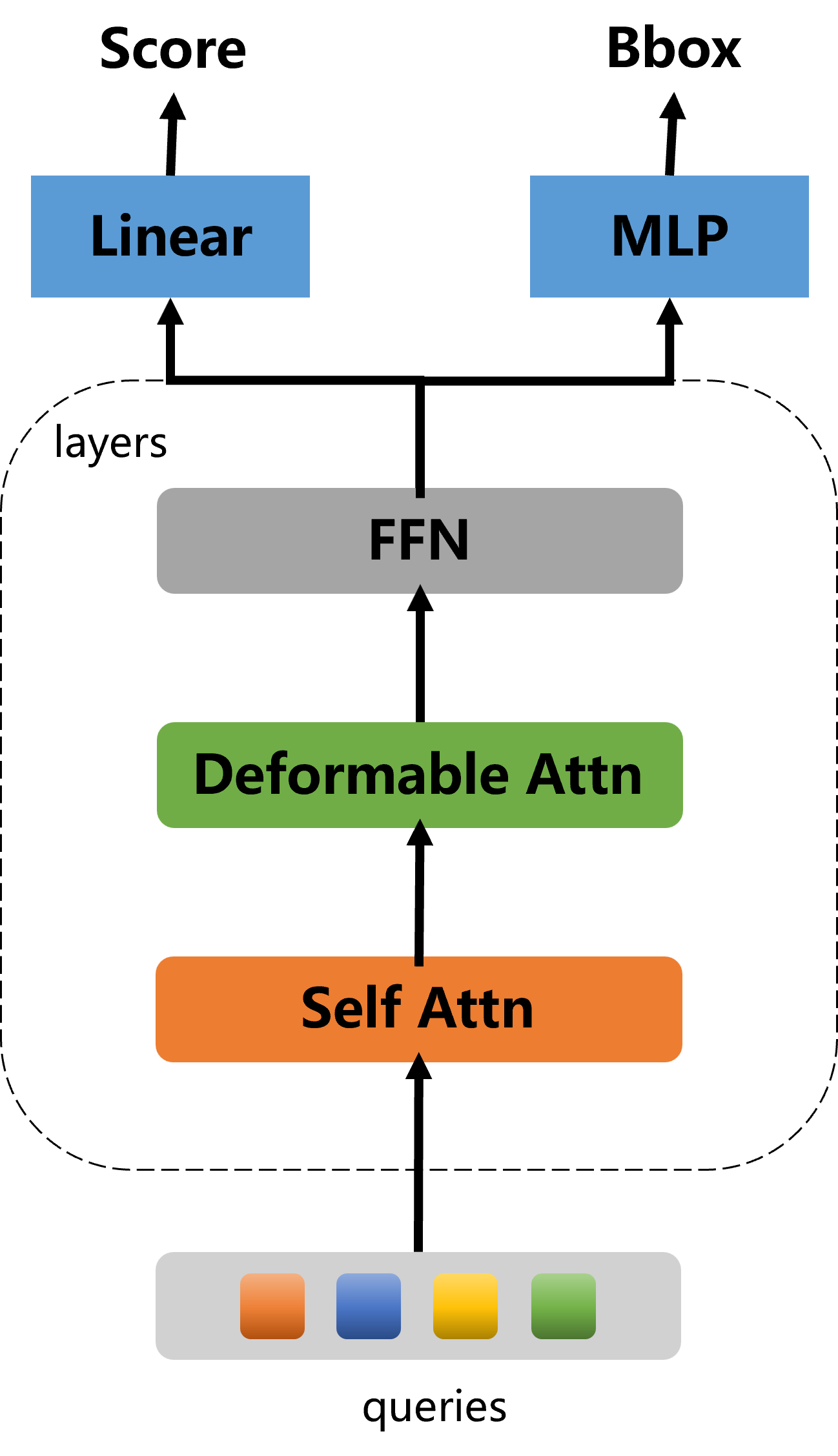}
\caption{Simplified architecture of a conventional DETR-like detection head, showcasing the typical stacking of self-attention and deformable cross-attention layers. Residual connections and normalization layers are omitted for clarity.}
\label{fig:detr_like_arch}
\end{figure}

\begin{figure}[h!]
\centering
\includegraphics[width=0.4\textwidth]{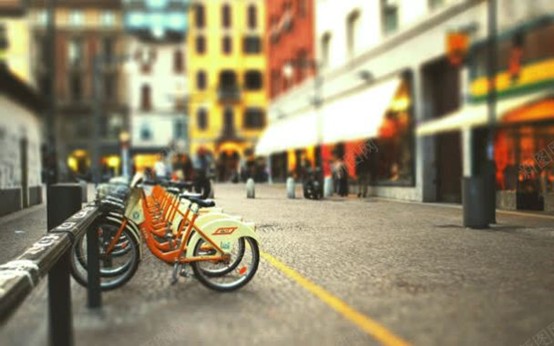}
\caption{Illustration simulating human visual focus. Identifying a bicycle's position and category does not necessitate sharp details of all surrounding elements like people or buildings; a somewhat blurred or general context often suffices.}
\label{fig:human_vision_analogy}
\end{figure}

We re-examined the design of these DETR-inspired detection heads, which typically involve repeatedly stacking self-attention and cross-attention layers (Figure~\ref{fig:detr_like_arch}). In DETR's decoder, for instance, a fixed set of learnable object queries interact with the entire image feature map via cross-attention to locate objects and then refine their predictions through self-attention among themselves. While this global interaction is powerful, our analysis revealed substantial redundancy, particularly in the self-attention modules: each object query, representing a potential object, performs pairwise attention with all other queries regardless of their spatial proximity or semantic relevance. This indiscriminate global attention often processes irrelevant information, increasing noise and computational burden. This design, while mathematically elegant, does not align intuitively with human visual perception: humans do not globally compare every potential object region to every other in a uniform manner; instead, we instinctively focus on key details, perceive the immediate surroundings, and glance at the broader context (Figure~\ref{fig:human_vision_analogy}). Inspired by this selective human attention, and recognizing the limitations of both fixed-parameter convolutions and overly global attention, we sought a more tailored attention mechanism. This involves restricting the range of attention to relevant local neighborhoods, similar to the locality of convolutions, but retaining the adaptive correlation-based computation inherent to attention. This approach allows for a more focused and efficient information aggregation, maximizing useful signals while significantly reducing noise from irrelevant interactions. While a naive limitation of receptive field might seem problematic for global context, we posit that a more efficient and human-interpretable attention mechanism, specifically designed for the detection head, could yield superior results by selectively attending to relevant information while retaining comprehensive contextual understanding.

To address these limitations and capture the efficiency of human visual processing, we propose the Context-Gated Scale-Adaptive Detection Network (CSDN). CSDN is a novel Transformer-based detection head inspired by our cognitive process of observing an object: first, concentrating on its precise location (Key Local Detail Mining); second, perceiving its immediate environment (Local Context Focus); and third, glancing at the broader context (Global Semantic Summary). Crucially, while this approach promotes focused attention, it does not inherently limit the overall receptive field. Instead, CSDN strategically and adaptively aggregates information from diverse scales and spatial relationships. It replaces a rigid stack of attention layers with a dynamic gating mechanism that allows each Region of Interest (ROI) to adaptively select and fuse features from three distinct attention patterns: a Global Semantic Summary (Block Attention, drawing top-level semantic features from FPN for macro background), a Local Context Focus (Neighbor Attention, based on IoU adjacency for immediate surroundings), and Key Local Detail Mining (Deformable Attention \cite{Dai2017ICCV} for precise discriminatory areas). This unique collaborative modeling of global and local features, dynamically weighted by the gating network, significantly streamlines the attention mechanism, reducing computational complexity by avoiding indiscriminate global interactions. Our proposed CSDN head is designed as a plug-and-play module, directly replacing the native heads of various CNN-based detectors (e.g., YOLO and ResNet-based models). With only minimal fine-tuning on pre-trained weights (24 epochs), CSDN consistently delivers significant accuracy improvements while maintaining an excellent balance of speed and precision, often outperforming conventional stacked Transformer heads.

Our main contributions are as follows: (i) We re-examine the necessity of exhaustive global interactions in self-attention mechanisms within DETR-like detection heads, demonstrating that restricting such interactions not only reduces redundancy but also leads to performance improvements through more focused and efficient attention. (ii) We introduce a gating network that adaptively aggregates three distinct Transformer-based attention patterns—global semantic summary, local context focus, and key local detail mining—resulting in more specific and interpretable feature channels that enhance contextual understanding. (iii) We develop a plug-and-play detection head module that seamlessly integrates with existing CNN-based detectors, achieving significant accuracy gains with minimal fine-tuning while maintaining real-time efficiency.

\newcommand{\YOLOvPlaceholder}[1]{YOLOv#1Placeholder} 

\section{Related Work}

\subsection{CNN-based Object Detectors: YOLO}
The YOLO series \cite{Redmon2016YouOL, Redmon2017YOLO9000BF, Redmon2018YOLOv3AI, Tian2025YOLOv12AC, \YOLOvPlaceholder{12}} exemplifies single-stage object detection, utilizing convolutional neural networks (CNNs) for both feature extraction and detection head. By processing an entire image in one pass, YOLO predicts bounding boxes and class probabilities directly, achieving remarkable inference speeds—e.g., YOLOv12\cite{\YOLOvPlaceholder{12}} reports a latency of 1.64 ms on T4 GPUs with a mean average precision (mAP) of 40.6\%. However, its detection head, built entirely on convolutional layers, struggles with precision in complex scenes due to limited capacity to model global context and long-range dependencies, often resulting in suboptimal performance for occlusions or multi-scale objects.

\subsection{Transformer-based Object Detectors: DETR and Variants}
DETR \cite{Carion2020EndToEndOD} introduced a Transformer-based detection head, redefining object detection as a set prediction task. It employs a CNN backbone for feature extraction, followed by a Transformer encoder-decoder stack as the detection head, using self-attention and cross-attention to predict bounding boxes and classes without anchor boxes or non-maximum suppression. Despite its elegance, DETR's reliance on multiple stacked Transformer layers leads to slow convergence and high computational costs. Variants like DINO \cite{Zhang2023DINO} enhance query mechanisms, while RT-DETR \cite{Li2023RTDETRRD} optimizes for real-time performance. Deformable DETR \cite{Zhu2021DeformableDD} refines this approach by incorporating a deformable attention mechanism in the head, reducing complexity by focusing on key points, yet these methods still depend on extensive layer stacking, posing efficiency challenges for real-time applications.

\subsection{Hybrid Approaches: YOLO and Transformers}
DEYO \cite{Ouyang2024DEYODW} combines a YOLO CNN backbone with a Transformer-based detection head, leveraging the strengths of both paradigms. Its head adopts a Transformer architecture to enhance contextual modeling beyond YOLO’s convolutional limitations, achieving higher accuracy than YOLO alone. DEYO employs a staged training strategy: the backbone is pre-trained with YOLO’s one-to-many label assignment, followed by fine-tuning of the Transformer head using a one-to-one matching approach. While this hybrid design outperforms YOLO, its inference speed does not rival RT-DETR, highlighting a trade-off between accuracy and real-time efficiency.

\subsection{Sparse Attention Mechanisms: Inspiration from NLP}

In natural language processing, sparse attention mechanisms like Native Sparse Attention (NSA) \cite{DeepSeek2025NativeSA} optimize efficiency by focusing on a subset of tokens rather than all pairwise interactions, effectively handling long sequences. Given that language is a one-dimensional sequence of information, we can extend this idea to images by viewing them as two-dimensional arrangements of 'words,' where each 'word' represents a pixel or region. This analogy suggests that attention mechanisms successful in NLP could be adapted to visual tasks by treating spatial arrangements of image regions similarly to token sequences.

\begin{figure}[h]
\centering
\includegraphics[width=1.0\linewidth]{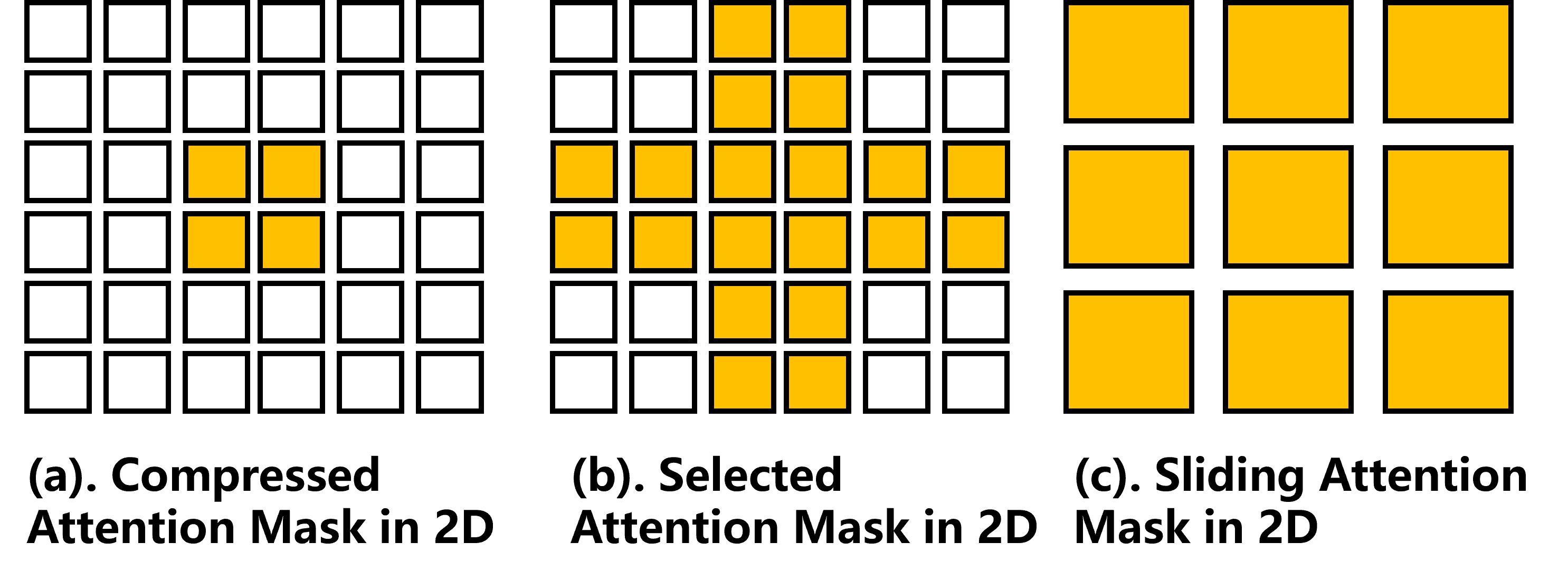}
\caption{2D extensions of NSA attention patterns, Yellow areas indicate regions where attention scores are computed, while white areas are skipped. Adapted from \cite{DeepSeek2025NativeSA}.}
\label{fig:nsa_2d}
\end{figure}

To illustrate this, consider placing the end of a sentence at the center of a 2D image, with surrounding 'words' as nearby pixels or regions. Adjacent areas naturally provide context, much like nearby words in a sentence, while semantically related regions—identified conveniently via Intersection over Union (IoU)—capture similar meanings despite spatial separation. Figure \ref{fig:nsa_2d} presents three 2D extensions of NSA attention patterns: Compressed Attention Mask, Selected Attention Mask, and Sliding Attention Mask. These diagrams, adapted from \cite{DeepSeek2025NativeSA}, show how attention can selectively focus on key regions (green) while skipping others (white), mirroring sparse token selection in language.

\section{The CSDN Model}

The Context-Gated Scale-Adaptive Detection Network (CSDN) introduces a Transformer-based detection head designed to enhance object detection by decoupling attention into three specialized modules: Block Attention, Neighbor Attention, and Deformable Attention. These modules emulate human visual cognition—global perception, local context analysis, and detailed observation, respectively—and are fused via a gated neural network for adaptive feature aggregation. This section outlines each component and the training methodology, leveraging insights from natural language processing and human vision to process image information efficiently.

\begin{figure}[h]
    \centering
    \includegraphics[width=0.8\linewidth]{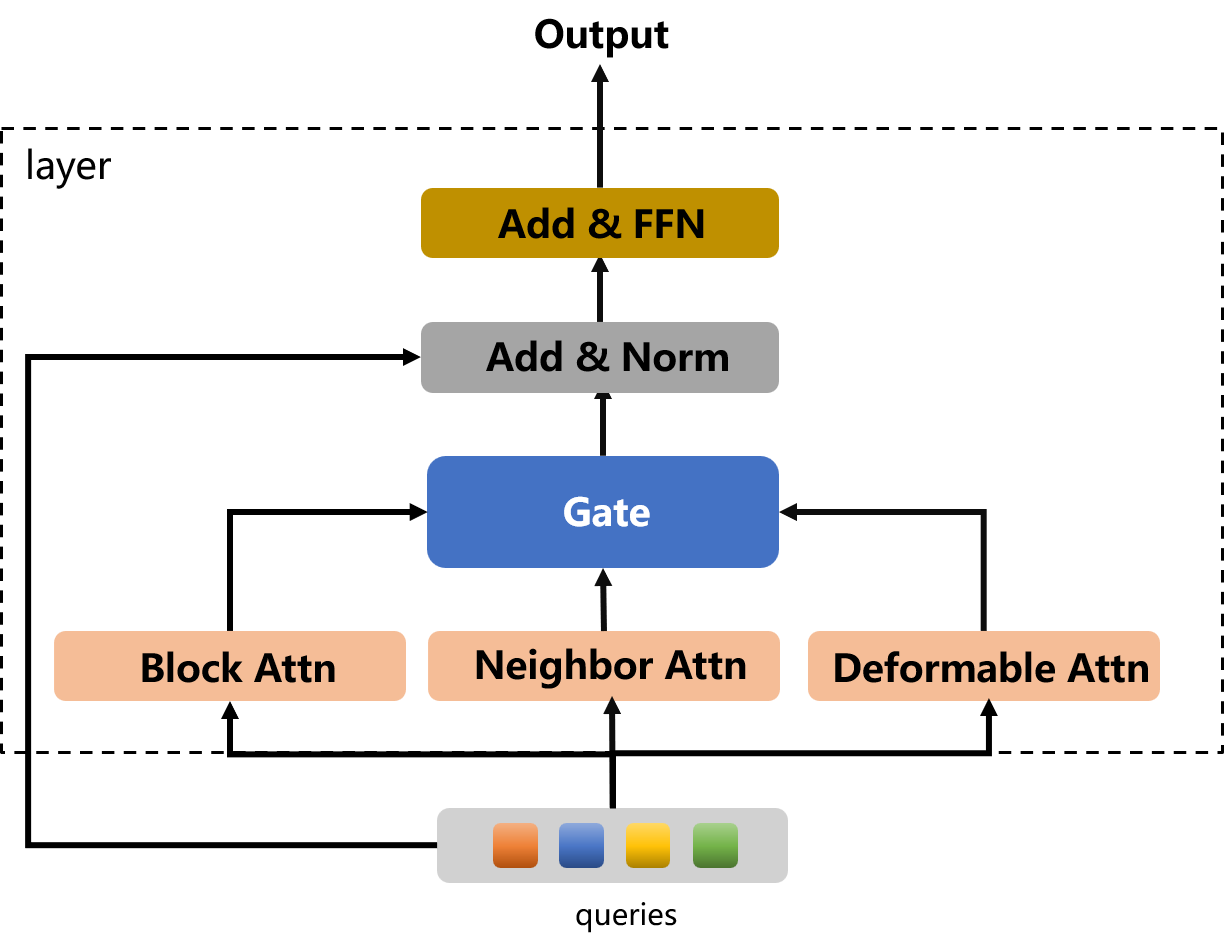}
    \caption{The complete architecture of CSDN, utilizing Gated Fusion to Integrate Outputs from Block, Neighbor, and Deformable Attention into later processing stages.}  
    \label{fig:gated_fusion}
\end{figure}

Each module targets a distinct aspect of contextual understanding. Block Attention captures global scene semantics, Neighbor Attention focuses on spatially relevant local interactions, and Deformable Attention extracts critical object details. The gated fusion mechanism dynamically integrates these outputs, as shown in Figure~\ref{fig:gated_fusion}, ensuring flexibility across diverse detection scenarios. This structured approach avoids the inefficiencies of traditional stacked attention, offering a balanced trade-off between computational cost and detection accuracy.

\subsection{Block Attention: Global Context Acquisition}

Block Attention simulates the human tendency to first grasp an overall scene impression before focusing on specific objects. Instead of exhaustively computing interactions across all image pixels, which is computationally prohibitive, this module leverages the top-level feature map \( F_{\text{top}} \) from the Feature Pyramid Network (FPN). \( F_{\text{top}} \) provides a compact, semantically rich summary of the entire image, enabling efficient global context integration.

\begin{figure}[h]
    \centering
    \includegraphics[width=0.8\linewidth]{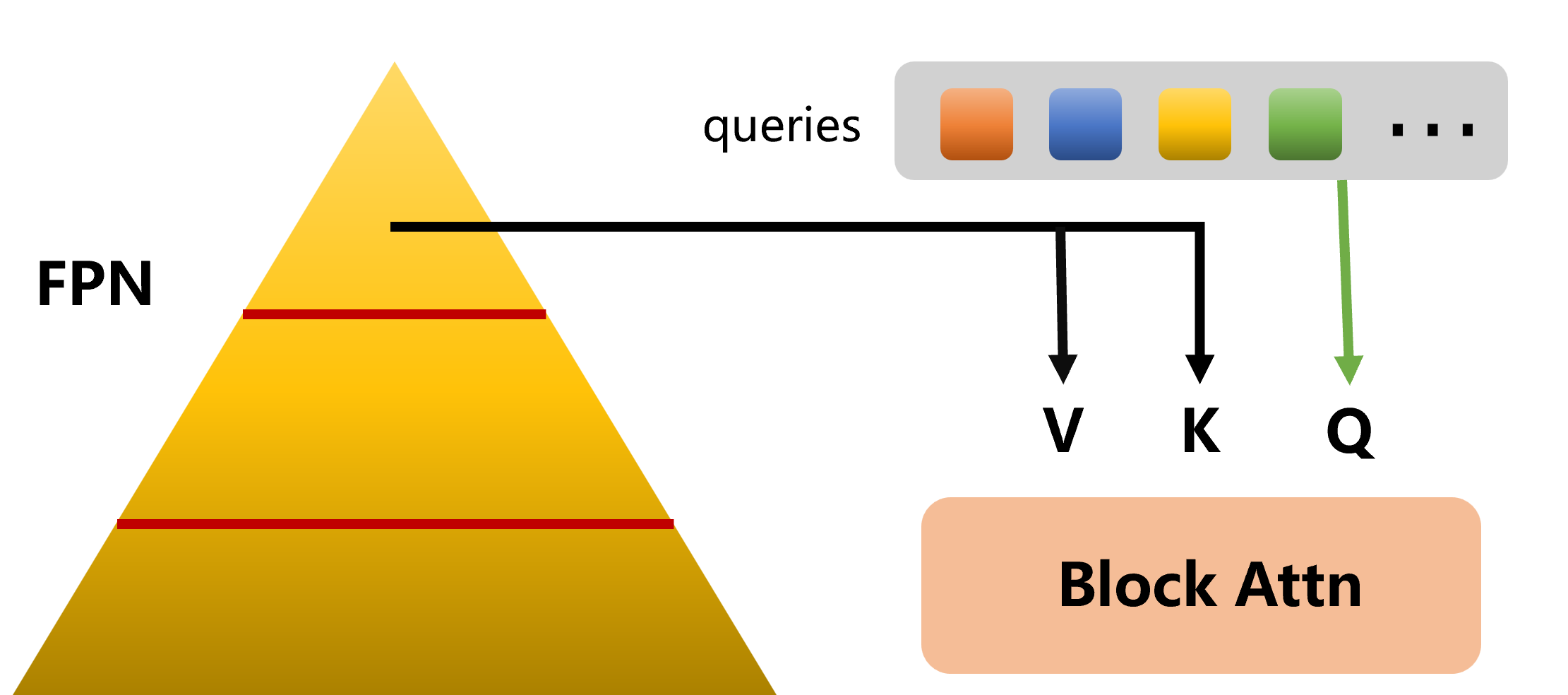}
    \caption{Block Attention mechanism. The input image’s multi-level FPN features are shown (left), with the Block Attention Module (right) computing cross-attention between queries and the top-level feature map \( F_{\text{top}} \) for global context.}
    \label{fig:block_attention}
\end{figure}

For each query \( q_i \) representing a region of interest, Block Attention performs cross-attention with \( F_{\text{top}} \), yielding an output \( O_b^{(i)} \) as:
\begin{equation}
    O_b^{(i)} = \text{softmax}\left( \frac{(q_i W_q) (F_{\text{top}} W_k)^T}{\sqrt{d_k}} \right) (F_{\text{top}} W_v),
\end{equation}
where \( W_q, W_k, W_v \) are learnable projections, and \( d_k \) scales the attention computation. This design, illustrated in Figure~\ref{fig:block_attention}, reduces complexity by focusing on compressed high-level features, equipping each query with a global perspective akin to recognizing a "street scene" before identifying its elements.

\subsection{Neighbor Attention: Local Context Refinement}

Once global context is established, Neighbor Attention refines understanding by focusing on spatially proximate regions, mirroring how humans examine an object’s immediate surroundings. Unlike traditional self-attention, which incurs high computational costs by interacting with all queries, this module restricts attention to neighboring queries with overlapping bounding boxes (IoU > 0), reducing noise from distant, irrelevant areas.

\begin{figure}[ht]
    \centering
    \includegraphics[width=0.8\linewidth]{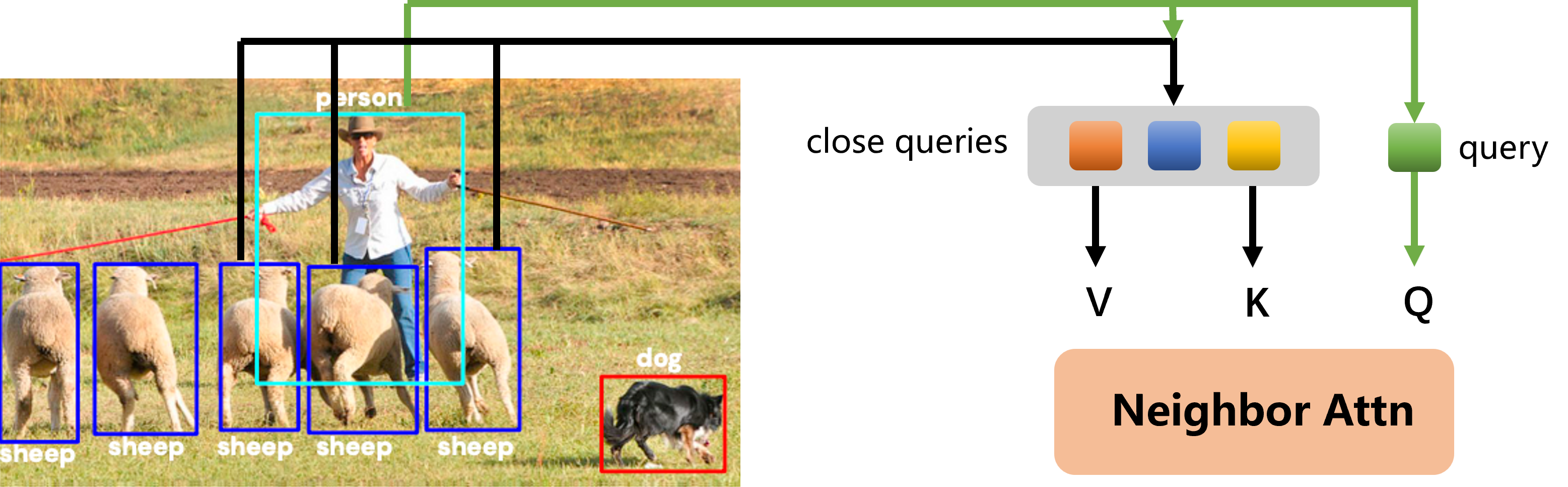}
    \caption{Neighbor Attention mechanism. The central query’s bounding box (green) interacts only with overlapping neighbors, ignoring distant regions, to refine local context.}
    \label{fig:neighbor_attention}
\end{figure}

For a query \( q_i \) with bounding box \( b_i \), its neighbor set \( \mathcal{N}_i \) includes queries \( q_j \) where \( \text{IoU}(b_i, b_j) > 0 \). The output \( O_n^{(i)} \) is:
\begin{equation}
    O_n^{(i)} = \sum_{j \in \mathcal{N}_i} \text{softmax}\left( \frac{(q_i W_q) (q_j W_k)^T}{\sqrt{d_k}} \right) (q_j W_v),
\end{equation}
where attention is masked for non-neighbors. Figure~\ref{fig:neighbor_attention} visualizes this selective focus, enhancing the model’s ability to discern spatially close or overlapping objects while maintaining efficiency, inspired by relational reasoning in networks like \cite{Hu2018RelationNetworks}.

\subsection{Deformable Attention: Detailed Feature Extraction}

\begin{figure}[h]
    \centering
    \includegraphics[width=0.8\linewidth]{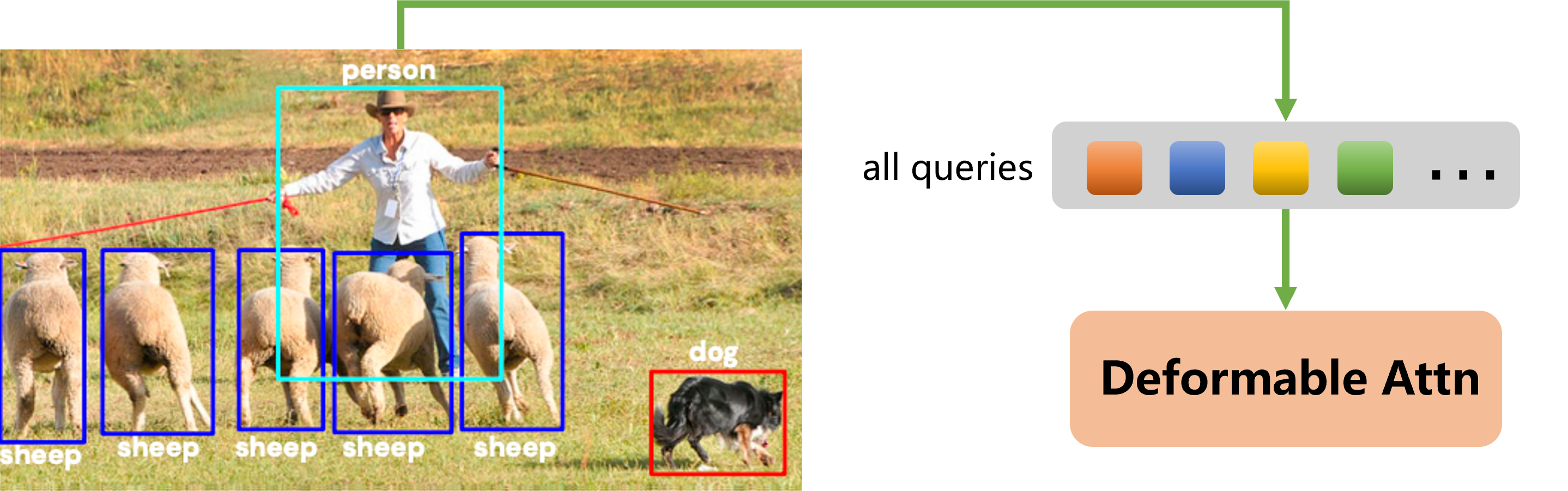}
    \caption{Deformable Attention mechanism. Learned offsets  adjust sampling points to focus on key area.}
    \label{fig:deformable_attention}
\end{figure}

Deformable Attention is incorporated to mimic the human visual system's ability to precisely focus on discriminative object parts, such as corners or texture changes, rather than uniformly scanning an entire area. Inspired by \cite{Zhu2021DeformableDD}, this mechanism enables the model to dynamically predict offsets for a small, learnable set of sampling points for each query. This adaptive sampling means the model can actively determine the most informative locations on the image feature map to extract features from, regardless of an object's irregular shape or varying pose. Such targeted sampling offers strong flexibility to adapt to diverse object characteristics, significantly enhances computational efficiency by avoiding redundant calculations over irrelevant regions, and ultimately yields higher accuracy in capturing fine-grained details critical for precise object identification and localization, as visualized in Figure~\ref{fig:deformable_attention}.

\subsection{Training Method}

To evaluate the CSDN detection head, we adopt a modular training strategy that retains a pre-trained CNN backbone (e.g., from COCO \cite{Lin2014ECCV}) and replaces the original head with CSDN, training it for 24 epochs on Titan V GPUs. This approach isolates the head’s performance, leverages established feature extraction, and enables efficient validation without retraining the entire network, facilitating integration into various architectures.

Training follows the RT-DETR framework \cite{Li2023RTDETRRD}, using Hungarian matching for prediction-ground truth assignment, a loss combining Focal Loss \cite{Lin2017FocalLF} for classification and L1 + GIoU \cite{Rezatofighi2019GeneralizedIO} for regression, and the AdamW optimizer \cite{Loshchilov2019DecoupledWD} with standard scheduling. Data augmentation includes random affine transformations and color jittering. Post-training, Non-Maximum Suppression (NMS) with a confidence threshold of 0.25 and IoU threshold of 0.6 refines predictions, ensuring robust performance comparisons with state-of-the-art models.

\newcommand{\LinCOCO}{Lin2014ECCV} 
\newcommand{\RTMDetPlaceholder}{RTMDetPlaceholder}
\newcommand{\RTDETRPlaceholder}{RTDETRPlaceholder}
\newcommand{\GoldYOLOPlaceholder}{GoldYOLOPlaceholder}
\newcommand{\DFinePlaceholder}{DFinePlaceholder}

\section{Experiments}

We evaluate the proposed Context-Gated Scale-Adaptive Detection Network (CSDN) on the COCO dataset \cite{Lin2014ECCV}, following standard protocols for object detection. All experiments use pre-trained backbones from YOLO models, with CSDN replacing the original detection head and fine-tuned for 24 epochs on Titan V GPUs. Metrics include precision (P), recall (R), mean average precision at IoU 0.5 (mAP50), and across 0.5 to 0.95 (mAP50-95 or AP). Below, we present ablation studies, performance analyses, and comparisons with state-of-the-art methods.

\subsection{Ablation Studies}

To dissect the contributions of CSDN's components, we conduct ablation experiments on the YOLOv8 backbone, focusing on attention mechanisms and their aggregation.

\subsubsection{Different Attention Mechanisms with Stacked Connections}

We first examine the necessity of global self-attention in DETR-style heads by comparing stacked configurations: (1) standard self-attention followed by deformable attention (s-d), (2) neighbor attention (IoU-based) followed by deformable attention (n-d), and (3) block attention followed by deformable attention (b-d). These variants test whether restricting interactions to local or global summaries improves over unrestricted global self-attention.

As shown in Table~\ref{tab:ablation_attention_types}, the n-d configuration outperforms the s-d baseline, achieving higher mAP50 (71.1 vs. 70.3) and mAP50-95 (54.9 vs. 54.5). This indicates that global self-attention may introduce noise from irrelevant distant queries, while IoU-constrained neighbor attention focuses on pertinent local contexts, reducing redundancy and enhancing precision. The b-d variant shows improved recall (64.0 vs. 63.1) but slightly lower precision, suggesting that cross-attention with FPN's top-level features provides a broader environmental context beneficial for detecting more objects, though it may dilute fine-grained accuracy due to the condensed representation. These results challenge the reliance on exhaustive global interactions, aligning with prior works on attention optimization in DETR variants \cite{Zhu2021DeformableDD}.

\begin{table}[h!]
\centering
\renewcommand{\arraystretch}{1.3}
\caption{Ablation on stacked attention mechanisms (- denotes stacking) using YOLOv8 backbone. s, n, b, d denote self, neighbor, block, and deformable attention, respectively.}
\label{tab:ablation_attention_types}
\begin{tabular}{c|cccc}
\toprule
Model & P    & R    & mAP50 & mAP50-95 \\ \midrule
s - d & 73.3 & 63.1 & 70.3  & 54.5     \\
n - d & 74.2 & 63.3 & 71.1  & 54.9     \\
b - d & 72.4 & 64.0 & 70.6  & 54.3     \\ \bottomrule
\end{tabular}
\end{table}

\subsubsection{Effectiveness of Multi-Attention Modal Gating Aggregation}

Building on the above, we assess the gated aggregation mechanism by fusing outputs from multiple attention types (+ denotes gating): self + deformable (s + d), block + deformable (b + d), neighbor + deformable (n + d), and all three (n + b + d). This evaluates whether adaptive weighting via the gated network effectively combines complementary information.

Table~\ref{tab:ablation_gated_aggregation} demonstrates that simple fusion of self and deformable attention (s + d) degrades performance (mAP50 62.8), likely due to conflicting global redundancies. In contrast, b + d maintains parity with its stacked counterpart, while n + d surpasses the baseline (mAP50 71.1 vs. 70.3), confirming the value of local focus. The full n + b + d aggregation yields the best results (mAP50 71.5, mAP50-95 55.1), highlighting the gated mechanism's ability to dynamically balance global (block), local (neighbor), and detailed (deformable) features. This adaptive fusion mitigates individual weaknesses, such as block attention's lower precision, leading to superior overall detection.

\begin{table}[h!]
\centering
\renewcommand{\arraystretch}{1.3}
\caption{Ablation on gated aggregation (+) of attention modules using YOLOv8 backbone, compared to s - d baseline.}
\label{tab:ablation_gated_aggregation}
\begin{tabular}{c|cccc}
\toprule
Model        & P    & R    & mAP50 & mAP50-95 \\ \midrule
s - d        & 73.3 & 63.1 & 70.3  & 54.5     \\
s + d        & 66.1 & 55.3 & 62.8  & 49.8     \\
b + d        & 71.5 & 63.0 & 70.0  & 54.3     \\
n + d        & 73.5 & 63.5 & 71.1  & 54.9     \\
\textbf{n + b + d}    & \textbf{74.3} & \textbf{64.2} & \textbf{71.5}  & \textbf{55.1}     \\ \bottomrule
\end{tabular}
\end{table}

\subsection{Performance Analysis}

We analyze CSDN's effectiveness as a plug-and-play head replacement and its sensitivity to layer depth.

First, we integrate CSDN into various YOLO models (YOLOv5x, YOLOv8x, YOLOv11x) and fine-tune for 24 epochs. Table~\ref{tab:compare_yolo_enhancement} shows consistent improvements: AP increases by 1.0 for YOLOv5x (53.4 to 54.4) and YOLOv8x (54.1 to 55.1), and 0.7 for YOLOv11x (54.9 to 55.6). These gains, achieved with minimal training, underscore CSDN's ability to enhance CNN backbones by better modeling contexts, boosting precision and mAP without significantly altering recall. This versatility suggests CSDN's broad applicability across evolving YOLO architectures.

\begin{table}[h!]
\centering
\renewcommand{\arraystretch}{1.3}
\caption{Performance improvements with CSDN as a replacement head for YOLO models. AP is mAP@[.5:.95].}
\label{tab:compare_yolo_enhancement}
\begin{tabular}{l|cccc}
\toprule
Method        & AP & AP50 & P    & R    \\ \midrule
YOLOv5x       & 53.4                 & 69.8 & 72.8 & 63.8 \\
\quad + CSDN  & \multicolumn{1}{c}{54.4 \textbf{($\uparrow$1.0)}} & \multicolumn{1}{c}{70.6} & \multicolumn{1}{c}{71.9} & \multicolumn{1}{c}{63.4} \\ \addlinespace
YOLOv8x       & 54.1                 & 70.7 & 73.5 & 64.7 \\
\quad + CSDN  & \multicolumn{1}{c}{55.1 \textbf{($\uparrow$1.0)}} & \multicolumn{1}{c}{71.5} & \multicolumn{1}{c}{74.3} & \multicolumn{1}{c}{64.2} \\ \addlinespace
YOLOv11x      & 54.9                 & 71.3 & 73.6 & 65.9 \\
\quad + CSDN  & \multicolumn{1}{c}{55.6 \textbf{($\uparrow$0.7)}} & \multicolumn{1}{c}{72.0} & \multicolumn{1}{c}{73.7} & \multicolumn{1}{c}{64.8} \\ \bottomrule
\end{tabular}
\end{table}

Next, we vary the number of Transformer layers in CSDN (2, 4, 6) using the YOLOv11x backbone and n + b + d aggregation. Table~\ref{tab:ablation_layer_depth} reveals that 4 layers achieve AP 55.6, matching 6 layers while reducing latency (23.2 ms vs. 23.6 ms) and outperforming 2 layers (AP 55.4). This indicates diminishing returns beyond 4 layers, with deeper stacks adding minor gains at the cost of efficiency. The 4-layer variant balances accuracy and speed, making it suitable for real-time applications, while we use 6 layers for SOTA comparisons to maximize competitiveness.

\begin{table}[h!]
\centering
\renewcommand{\arraystretch}{1.3}
\caption{Performance with varying CSDN layers on YOLOv11x. AP is mAP@[.5:.95].}
\label{tab:ablation_layer_depth}
\begin{tabular}{c|ccccc}
\toprule
Layers & P    & R    & mAP50 & AP   & Latency(ms) \\ \midrule
2      & 74.5 & 63.8 & 71.8  & 55.4 & 21.7        \\
4      & 74.1 & 64.3 & 71.9  & 55.6 & 23.2        \\
6      & 73.7 & 64.8 & 72.0  & 55.6 & 23.6        \\ \bottomrule
\end{tabular}
\end{table}

\subsection{Comparison with State-of-the-Art Methods}

We compare YOLOv11x + CSDN (6 layers) against leading real-time detectors on COCO. Table~\ref{tab:compare_sota_final} shows our model achieves AP 55.6 and AP50 72.0, surpassing YOLOv10x (AP 54.5) \cite{Tian2025YOLOv12AC}, baseline YOLOv11x (AP 54.9), and RT-DETR-HGNetv2-X (AP 54.8) \cite{Li2023RTDETRRD}, while being competitive with YOLOv12x (AP 55.2) and D-Fine (AP 55.8). With 63.4M parameters and 194.1 GFLOPs, it offers a strong efficiency-accuracy trade-off (23.6 ms latency), outperforming heavier models like RTMDet-X (AP 52.8) \cite{\RTMDetPlaceholder}. These results affirm CSDN's efficacy in elevating existing detectors to SOTA levels without paradigm shifts.

\begin{table*}[h!]
\centering
\renewcommand{\arraystretch}{1.3}
\caption{YOLOv11x + Ours vs. Other Real-time Object Detectors on COCO (using the largest scale for each model without pre-trained on Object360). AP refers to mAP@[.5:.95].}
\label{tab:compare_sota_final}
\begin{tabular}{ll|ccccc}
\toprule
Category & Method & AP & AP50 & Params(M) & GFLOPs & Latency(ms) \\ \midrule
\textit{YOLO Series} & YOLOv10x \cite{\YOLOvPlaceholder{10}} & 54.5 & 71.0 & 31.7 & 170.6 & 14.8 \\
& YOLOv11x (Baseline) \cite{\YOLOvPlaceholder{11}} & 54.9 & 71.3 & 56.9 & 194.9 & 17.5 \\
& YOLOv12x \cite{\YOLOvPlaceholder{12}} & 55.2 & 72.0 & 59.1 & 199   & /    \\
& RTMDet-X \cite{\RTMDetPlaceholder} & 52.8 & 70.4 & 94.9 & 141.7 & /    \\
& GOLD-YOLO-L \cite{\GoldYOLOPlaceholder} & 52.3 & 69.6 & 75.1 & 151.7 & /    \\ \addlinespace
\textit{DETR Series} & RT-DETR-HGNetv2-X \cite{\RTDETRPlaceholder} & 54.8 & 73.1 & 67   & 234   & 20.6 \\
& D-Fine \cite{\DFinePlaceholder} & 55.8 & 73.7 & 62   & 202   & /    \\ \addlinespace
\textbf{Ours} & \textbf{YOLOv11x + CSDN} & \textbf{55.6} & \textbf{72.0} & \textbf{63.4} & \textbf{194.1} & \textbf{23.6} \\ \bottomrule
\end{tabular}
\end{table*}

\section{Conclusion and Outlook}
\label{sec:conclusion}

In this work, we present the Context-Gated Scale-Adaptive Detection Network (CSDN), a Transformer-based detection head that addresses limitations in CNN receptive fields and inefficiencies in traditional DETR-like self-attention. By decoupling attention into specialized modules—block for global context, neighbor for local interactions, and deformable for detailed features—and fusing them via a gating mechanism, CSDN enables adaptive, scale-aware context modeling. Extensive experiments on COCO demonstrate its efficacy: ablations reveal redundancy in global self-attention and the superiority of gated aggregation over stacking, while CSDN enhances YOLO models (e.g., +1.0 AP on YOLOv8x) and achieves competitive performance with fewer layers, balancing accuracy and efficiency.

Future research can refine CSDN by specializing attention modules, such as restricting deformable attention to lower FPN levels for focused detail extraction and optimizing block attention with fewer representative global features to reduce overhead. Additionally, exploring alternative neighbor definitions beyond IoU, such as semantic similarity or graph-based methods, could further improve local context refinement. We envision the CSDN head's adaptive, decoupled design finding broader applications in diverse detection frameworks, enhancing real-time systems across computer vision tasks.

{
    \small
    \bibliographystyle{ieeenat_fullname}
    \bibliography{thesis}
}


\end{document}